\documentclass[letterpaper, 10 pt, conference]{ieeeconf}  

\IEEEoverridecommandlockouts                              

\usepackage{graphicx}%
\usepackage{multirow}%
\usepackage{array}
\usepackage{amsmath,amssymb,amsfonts}%
\usepackage{mathrsfs}
\usepackage{dblfloatfix}
\usepackage{cite}
\usepackage{caption}
\usepackage{subcaption}
\usepackage{color,soul}
\usepackage{pdfpages}
\usepackage{algpseudocode}
\usepackage{booktabs}
\usepackage{float}
\usepackage[ruled, vlined]{algorithm2e}

\title{\LARGE \bf
Human-in-the-loop Optimisation in Robot-assisted Gait Training
}

\author{Andreas Christou$^{1}$, Andreas Sochopoulos$^{1}$, Elliot Lister$^{1}$ and Sethu Vijayakumar$^{1}$
\thanks{*This research was supported in part by the Engineering and Physical Sciences Research Council (EPSRC, grant reference EP/L016834/1) as part of the Centre for Doctoral Training in Robotics and Autonomous Systems at Heriot-Watt University and The University of Edinburgh, in part by the Japanese Science and Technology Agency through the Moonshot project, and in part by the Alan Turing Institute, UK, and by the Honda Research Institute Europe.}
\thanks{$^{1}$Andreas Christou, Andreas Sochopoulos, Elliot Lister and Sethu Vijayakumar are with the School of Informatics, University of Edinburgh, UK.
{\tt\small andreas.christou@ed.ac.uk}}%
}

\begin{document}

\maketitle
\thispagestyle{empty}
\pagestyle{empty}

\begin{abstract}
Wearable robots offer a promising solution for quantitatively monitoring gait and providing systematic, adaptive assistance to promote patient independence and improve gait. However, due to significant interpersonal and intrapersonal variability in walking patterns, it is important to design robot controllers that can adapt to the unique characteristics of each individual. This paper investigates the potential of human-in-the-loop optimisation (HILO) to deliver personalised assistance in gait training. The Covariance Matrix Adaptation Evolution Strategy (CMA-ES) was employed to continuously optimise an assist-as-needed controller of a lower-limb exoskeleton. Six healthy individuals participated over a two-day experiment. Our results suggest that while the CMA-ES appears to converge to a unique set of stiffnesses for each individual, no measurable impact on the subjects' performance was observed during the validation trials.
These findings highlight the impact of human-robot co-adaptation and human behaviour variability, whose effect may be greater than potential benefits of personalising rule-based assistive controllers. Our work contributes to understanding the limitations of current personalisation approaches in exoskeleton-assisted gait rehabilitation and identifies key challenges for effective implementation of human-in-the-loop optimisation in this domain.
\end{abstract}


\section{INTRODUCTION}

Wearable robotic devices hold great promise in enhancing the outcomes of physical therapy and reducing the physical strain on healthcare professionals. However, the growing use of robotic assistance introduces the risk of depersonalising rehabilitation, potentially losing the tailored, highly effective treatments typically provided by healthcare providers \cite{Pennycott2012d,Israel2007c}. To address this concern, there has been a shift toward the use of collaborative robots designed to cater to the specific needs of patients, offering personalised assistance.

In gait rehabilitation, several studies have emphasised the importance of providing ``assistance as needed"—partial support that encourages the patient's active participation \cite{Pennycott2012d,Paolucci2012,Kaelin-Lane2005,Lotze2003}. Although various control strategies exist to achieve this \cite{Shahriari2019, Taherifar2018a, Hussain2017a, Russi2016c, Duschau-Wicke2010a}, it remains unclear whether there is a single control
strategy that is superior to other strategies or that is in its form optimal for each individual. A prevailing issue with current practices is that many existing controllers are tuned based on the performance of a healthy participant, resulting in generalised solutions that often fail to meet individual needs effectively. Given the significant variability in gait among individuals, it is crucial to explore new methods for adjusting control parameters in wearable devices to deliver personalised and adaptive assistance.

\begin{figure*}[h]
    \centering
   \includegraphics[width=0.85\linewidth]{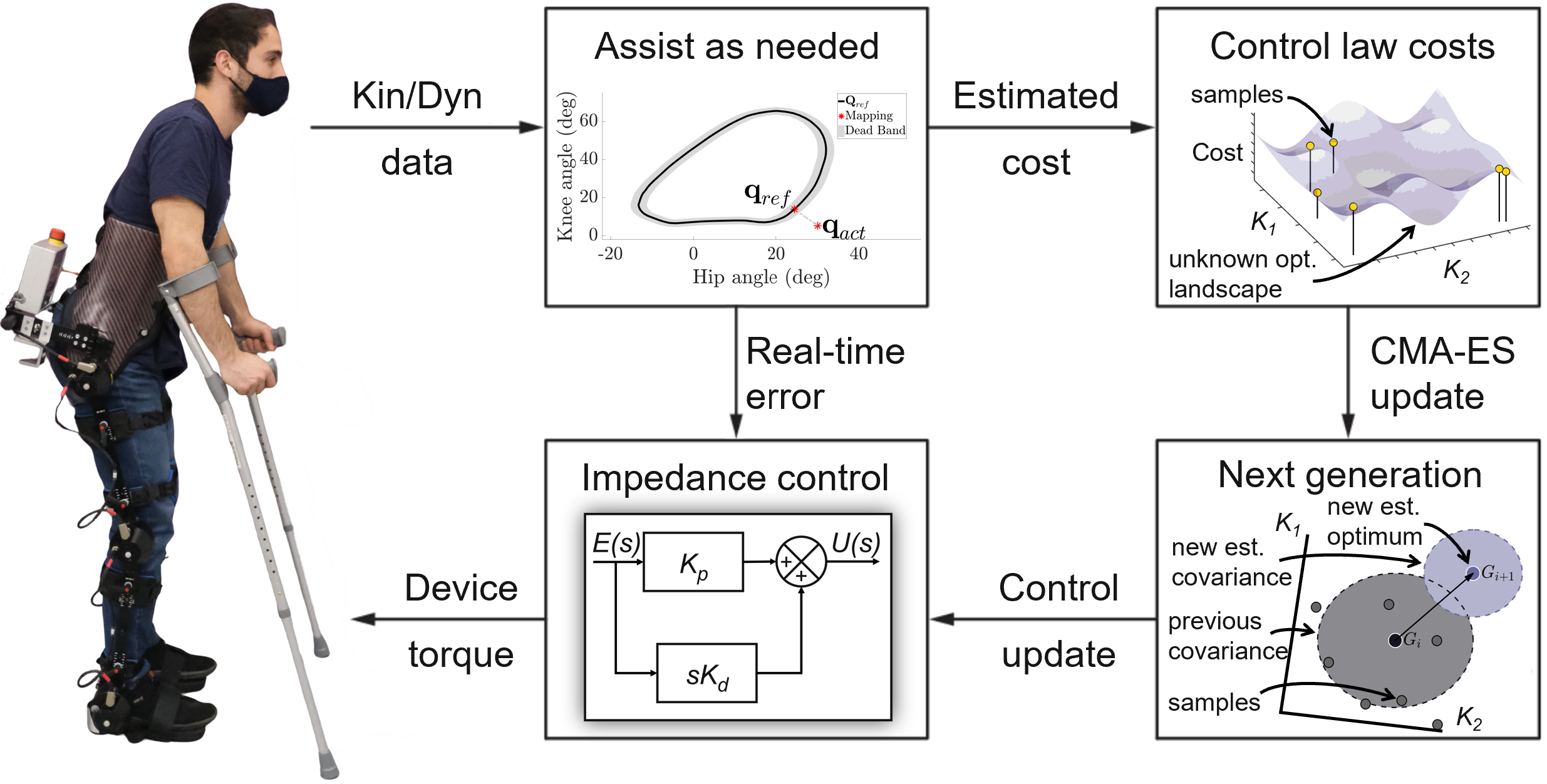}
    \caption{HILO pipeline using the CMA-ES to personalise the open parameters of an impedance controller to provide assistance as needed.}
    \label{fig:HILO_Zhang2017}
\end{figure*}

Human-in-the-loop optimisation (HILO) is a dynamic approach that integrates human feedback directly into the control system of wearable devices allowing for real-time adjustments and ensuring that the device adapts to the user's unique biomechanics and physiological responses. This method leverages the cyclic nature of gait, where adjustments to robot controllers are made iteratively with each gait cycle or number of gait cycles (Figure \ref{fig:HILO_Zhang2017}a). Based on this continuous feedback loop, HILO tailors robotic controllers to the needs of the user providing individualised assistance. 

To date, HILO has been successful in adjusting the assistance provided by wearable robots in order to reduce the metabolic cost of gait \cite{Gordon2022,Zhang2017,Poggensee2021,Bryan2021}, increase the self-selected speed of walking \cite{Song2021,Slade2022} in healthy subjects and reduce joint loading in manual material handling activities \cite{sochopoulos2023human}. Using mostly one-DOF robots, studies have focused on the optimisation of parameterised assistive torque profiles for mainly the hip joint and the ankle joint. However, the effectiveness of HILO for the personalisation of rehabilitation controllers has not been studied. It is hypothesised here that HILO could be used for the personalisation of the open parameters of rehabilitation controllers in order to provide assistance as needed.

So far, to carry out HILO, two main algorithms have been used: the Covariance Matrix Adaptation Evolution Strategy (CMA-ES) and Bayesian optimisation. Both methods, are sample-based derivative-free optimisation methods that search for the global optimum within a constrained space. However, their underlying assumptions differ which will likely affect the efficacy of HILO, particularly when applied to gait training where time-dependent gait variability is expected to be higher. 

In Bayesian optimisation, a surrogate model of a continuous function is constructed based on the sampled observations. Using Gaussian process regression, the posterior probability distribution of the unknown function, $f(x)$, is iteratively updated, and is used to update the acquisition function, $a(x)$. The updated acquisition function is then used to compute the next best sample point, $x_{i}$, and this process repeats. After $N$ observations, this process terminates and the point where the value of $f(x)$ is highest (or lowest) is obtained. To do this, a common acquisition function is the \textit{expected improvement} function, which balances exploration and exploitation. Using the \textit{expected improvement} acquisition function the next best sample point is computed as the point with the highest expected quality and the highest posterior standard deviation, based on the assumption that $f(x)$ (given observations $y_{1:n}$ at points $x_{1:n}$) is normally distributed \cite{Frazier2018}. This implementation assumes noise-free evaluations and a constant function $f$. However, when it comes to HILO, where $f$ represents the response of humans to an external force, some of these assumptions may be violated. It is important when using Bayesian optimisation to incorporate methods for updating the acquisition function based on realistic values for expected noise, and account for time-dependent changes in human behaviour due to fatigue, concentration and/or motor learning, which are often hard to model and predict.

In CMA-ES a stochastic search for the global optimum of an unknown function, $f$, is pursued iteratively through a series of generations, $g$. On every generation, a total of $\lambda$ samples, $\{x_k^g | k\in\mathbb{N}, 1\le k\le\lambda\}$, are generated by sampling a multivariate normal distribution with mean, ${\bf m}^{g}$, and covariance, ${\bf C}^{g}$, and are evaluated. Based on the observations, the new mean of the search distribution, ${\bf m}^{g+1} \in \mathbb{R}^{n}$, the new covariance matrix, ${\bf C}^{g+1} \in \mathbb{R}^{n \times n}$, and the new step size, $\sigma^{g+1} \in \mathbb{R}_{>0}$, are updated, where $n$ is the dimension of the search space. A step of size, $\sigma$, in a direction dictated by the sampled observations is performed, a new sample population is generated around the new mean and this process is repeated for $G$ generations. This process does not assume that the unknown function, $f$, is constant and does not prevent resampling of the same points. This allows time-dependent changes in function $f$ to be captured and noise in the sampled observations to be accounted for. 

In this work, we propose the personalisation of a lower-limb assist-as-needed controller using HILO and the CMA-ES. Through an experimental study, we observe the ability of HILO to adjust the open parameters of a gait training controller in order to help the users accurately follow a predefined kinematic path with minimal assistance. A continuous optimisation protocol is followed over a multi-day trial as described in \cite{Poggensee2021}. The results obtained from six healthy subjects are presented and discussed.

\section{HILO and Assistance As Needed}

\begin{figure}[h]
  \centering
  \includegraphics[width=0.95\linewidth]{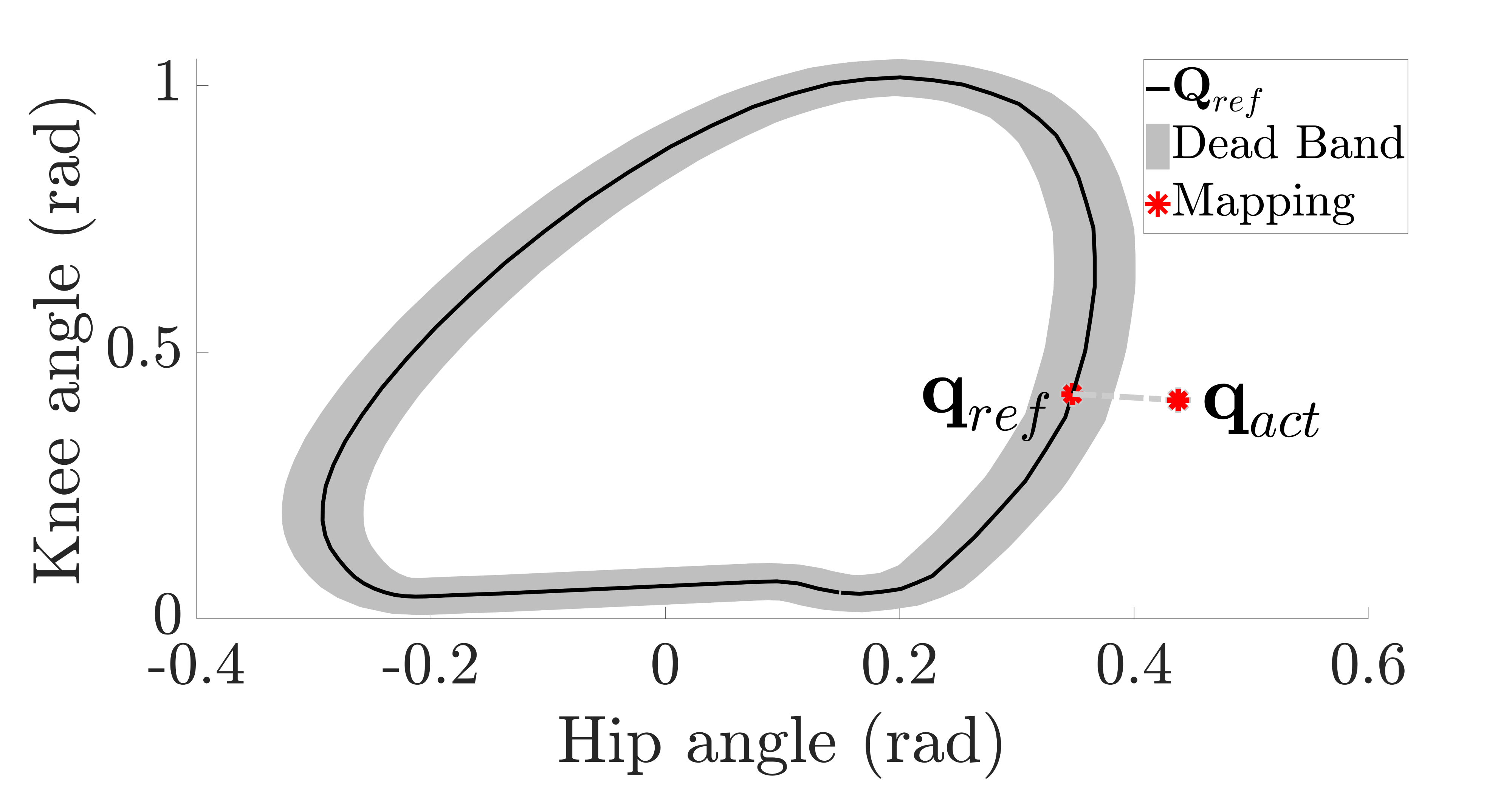}
  \caption{Illustration of the reference kinematic path, ${\bf Q}_{\text{ref}}$, surrounded by a dead band and the mapping of the kinematic configuration of the model, ${\bf q}_{\text{act}}$, to the reference point, ${\bf q}_{\text{ref}}$, on the reference path.}
  \label{fig:RefGaitCycle}
\end{figure}

Here we optimise the stiffness of an impedance controller using HILO in order to provide assistance as needed. Following the principles of path control  \cite{Duschau-Wicke2010a}, a reference kinematic path,  ${\bf Q}_{ref} \in \mathbb{R}^{i \times 2}$, is prescribed\footnote{This reference path is used as a means of verifying the proposed HILO. It is not implied that this path is ideal for all participants. 
} (where $i$ is the number of points in the discretised domain of the reference path), a dead band, $r_{db}$, is defined around the reference path and an impedance controller is used to ensure that the user's trajectory, ${\bf q}_{\text{act}} \in \mathbb{R}^{2}$, stays close to the reference (Figure \ref{fig:RefGaitCycle}). Iteratively, the stiffness of the hip and the knee joints of the two legs is then adjusted and a measure of the objective function value is obtained. With the aim to provide assistance as needed, the objective function is defined as:
\begin{align}
  \min_{\bf K} \quad \frac{w_{1}}{J_{1}}\frac{{\sum_{i=1}^{N-1} ( {\bf u}_{{r}_{i}}^{T}{\bf u}_{{r}_{i}})}}{N-1} &+ \frac{w_{2}}{J_{2}}\frac{\sum_{i=1}^{N}({{\Delta {\bf q}}}_{i}^{T}{{\Delta {\bf q}}}_{i})}{N} \notag\\ &+ \frac{w_{3}}{J_{3}}{\sum_{j=1}^{4}{K}_{j}},
  \label{eq:ObjFunHILO}
\end{align}
where the first cost aims to minimise the effort of the robot, the second aims to minimise the tracking error of the participant, and the third aims to prioritise solutions with a lower exoskeleton stiffness. The control effort cost aims to tune the assistive controller so that the exoskeleton intervenes as little as possible to the human movement, and is conflicting with the tracking cost, which aims to align human gait with the desired gait pattern. The stiffness cost acts as a regulariser, restricting the choice of unnecessarily large stiffness values that can lead to user discomfort.

The decision variables, ${\bf K}\in \mathbb{R}^4$, of the optimisation problem denote the stiffness for the hip and the knee joints, ${\bf u}_{r}$ is the exoskeleton assistance, ${\Delta {\bf q}}$ is the kinematic tracking error, and $N$ is the number of time steps recorded in one observation. ${\bf w}$ is the vector of normalised weights for the three costs, and $J$ is a vector of scaling factors. The scaling factors are used to normalise the cost terms to the maximum exoskeleton assistance (${\bf u}_{max}=40$Nm), the maximum expected trajectory error ($\Delta {\bf q}_{max}=2$\textdegree), and the maximum expected stiffness (${\bf K}_{max}=400$Nm/rad), respectively, such that the magnitude of the costs is comparable. Similarly, $N$ is used to normalise the costs to the length of the recorded time steps and $w$ is used to adjust the relative importance of the normalised costs. For this study, the weights in the objective function were heuristically defined to prioritise tracking performance and keep the contribution of the third cost (which favours solutions with low stiffness) to approximately 10\% of the objective function value (${\bf w}=[3, 1, 0.1]$). 

Based on path control, the kinematic tracking error, ${\Delta {\bf q}}$, and the exoskeleton assistance, ${\bf u}_{r}$, are defined as:
\begin{gather}
    {\Delta{\bf \tilde q}} = {{\bf q}}_{\text {ref}} - {\bf q}_{\text {act}}, \label{eq:pathControl1} \\
    {\Delta {q}}^{(j)} =
    \begin{cases}
        0 , & |{\Delta \tilde q}^{(j)}| \leq r_{\text {db}}, \\
        {\Delta \tilde q}^{(j)} - r_{\text {db}} , & {\Delta \tilde q}^{(j)} > r_{\text {db}},  \\
        {\Delta \tilde q}^{(j)} + r_{\text {db}} , & {\Delta \tilde q}^{(j)} < -r_{\text {db}},
    \end{cases}\\
    {\boldsymbol u}_{r} = {\bf K}{\Delta{{\bf q}}} + {\bf B}{\Delta{{\dot{\bf q}}}}, \label{eq:impedanceController1} \\
    {\bf B}={\bf c}_{cr}\sqrt{\bf K}, \label{eq:impedanceController2}
\end{gather}
where ${\bf q}_{\text{ref}} \in \mathbb{R}^{2}$ is the reference point that is dynamically defined as the point on the reference path that is geometrically closest to the pose of the human, and $\bf B$ and ${\bf c}_{cr}$ are the matrix of joint damping and the matrix of the critical damping coefficients of the exoskeleton, respectively. 

\section{METHODOLOGY}
To speed up the convergence of the optimisation through an efficient sampling method, the CMA-ES is used as described in \cite{Hansen2016}. The mean value of the search distribution at the first generation is defined, ${\bf m}^{0}$, and $\lambda$ sample points, $\{x_{k}^{1}| k \in \mathbb{N}, 1 \le k \le \lambda\}$ are generated based on a multivariate normal distribution with zero mean and unit variance, ${\bf C}^{0}=I$. The performance of the subjects at the generated sample points is calculated using equation \ref{eq:ObjFunHILO} and the value of the mean point is updated based on a number of the sampled points, $\{\mu | \mu < \lambda\}$, which are weighted according to the subject's performance. This can be expressed as \cite{Hansen2016}:
\begin{align}
  {\bf x}^{g+1}_k \sim {\bf m}^{g} +  \sigma^{g}\mathcal{N}(0,{\bf C}^{g}) \ \text{for} \ k=1,2,...,\lambda,\\
  {\bf m}^{g+1} = {\bf m}^{g} + c_{m}\sigma^{g}\sum_{i=1}^{\mu}w_{i}({\bf x}_{i:\lambda}^{g+1}-{\bf m}^{g}),
\end{align}
where the symbol $\sim$ denotes the same distribution on the left and right side, $\sigma$ is the step size, $c_{m}$ is the learning rate for the mean and ${\bf x}_{i:\lambda}^{g+1}$ is the $i$-th best sample out of all samples, ${\bf x}_k^{g+1}$, with the index $i:\lambda$ denoting the index of the $i$-th ranked sample such that $f({x}_{1:\lambda}^{g+1}) \le f({x}_{2:\lambda}^{g+1}) \le ... \le f({x}_{\lambda:\lambda}^{g+1})$, where $f$ is the objective function to be minimised. 

\begin{figure*}[t!]
  \centering
  \includegraphics[width=0.93\linewidth]{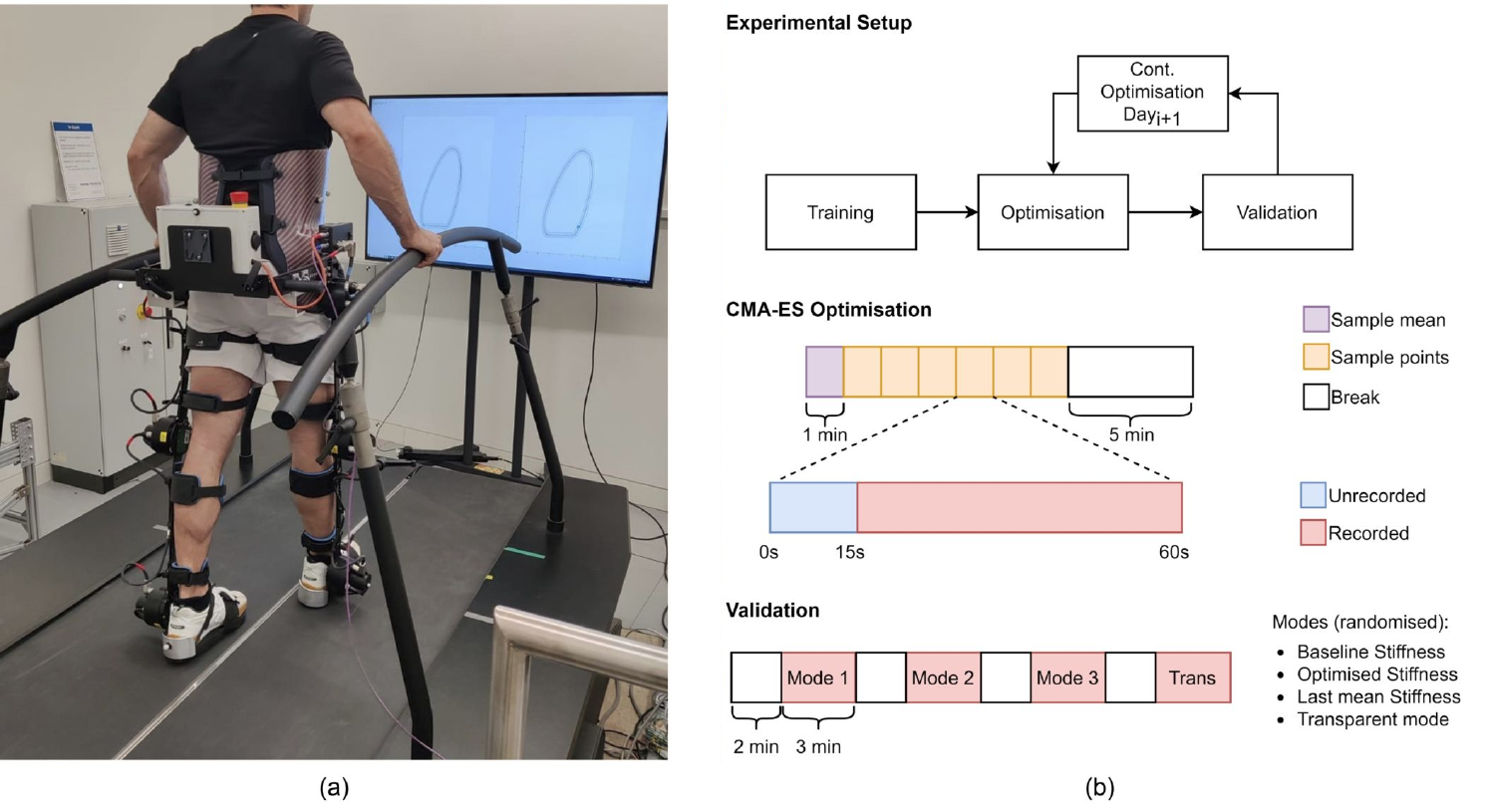}
  \caption[HIL optimisation experiments]{(a) Healthy participant walking on a self-paced treadmill with real-time visual feedback and assistance from the exoskeleton, Exo-H3. (b) Experimental protocol for HILO following a continuous optimisation protocol over multiple days.}
  \label{fig:HILOExperiments}
\end{figure*}
On every generation, an update of the covariance matrix, ${\bf C}^{g}$, and the step size, $\sigma^{g}$, is carried out. The covariance matrix is updated such that it retains information from both the entire population, and the correlations between generations. To do this, the \textit{cumulative} evolution path, ${\bf p}_{c}^{g+1}$, is utilised, which is the sequence of steps CMA-ES takes over a number of generations. This is expressed as \cite{Hansen2016}:
\begin{gather}
    \begin{split}
  {\bf C}^{g+1} = (1+c_{1}\delta(h_{\sigma})-c_{1}-c_{\mu}){\bf C}^{g}+c_{1}{\bf p}_{c}^{g+1}{{\bf p}_{c}^{g+1}}^{T} \\ + c_{\mu}\sum_{i=1}^{\mu}w_{i}{\bf y}_{i:\lambda}^{g+1}{{\bf y}_{i:\lambda}^{g+1}}^{T},
  \end{split}\\
  {\bf p}_{c}^{g+1} = (1-c_{c}){\bf p}_{c}^{g} + h_{\sigma}\sqrt{c_{c}(2-c_{c})\mu_{eff}}\sum_{i=1}^{\mu}w_{i}{\bf y}_{i:\lambda}^{g+1},\\
  {\bf y}_{i:\lambda}^{g+1} = ({\bf x}_{i:\lambda}^{g+1} - {\bf m}^{g})/{\sigma}^{g},
\end{gather}
where $c_{1}$, $c_{\mu}$ and $c_c$ are the learning rates for the rank-\textit{one} and rank-$\mu$ updates of the covariance matrix and the \textit{cumulative} evolution path, respectively, $\mu_{eff}$ is the effective sample size of the selected samples defined as $\mu_{eff}=1/{\sum_{i=1}^{\mu}w_i^2}$, $\delta(h_{\sigma})$ is defined as ${\delta}(h_{\sigma})=(1-h_{\sigma})c_{c}(2-c_{c}) \le 1$, and $h_{\sigma}$ is a Heaviside function that stalls the update of the \textit{cumulative} evolution path depending on the size of the \textit{conjugate} evolution path. The \textit{conjugate} evolution path, ${\bf p}_{\sigma}^{g}$, is independent of the direction of the successive steps performed, and is used to update the step size. The function, $h_{\sigma}$, and the \textit{conjugate} evolution path are defined as \cite{Hansen2016}:
\begin{gather}
  h_{\sigma} =
  \begin{cases}
    1, &\text{if} \frac{||{\bf p}_{\sigma}^{g+1}||}{\sqrt{1-(1-c_{\sigma}^{2(g+1)})}} < (1.4 + \frac{2}{n+1}) \mathbb{E} ||\mathcal{N}({\boldsymbol{0}},{\bf I})||, \\
    0, & \text{otherwise}, \\
  \end{cases}\\
  \mathbb{E}||\mathcal{N}({\boldsymbol{0}},{\bf I})|| \approx \sqrt{n}(1-\frac{1}{4n}+\frac{1}{21n^{2}}),
  \end{gather}
\begin{gather}  
  {\bf p}_{\sigma}^{g+1} = (1-c_{\sigma}){\bf p}_{\sigma}^{g}+\sqrt{c_{\sigma}(2-c_{\sigma}){\mu_{eff}}}{\bf C}^{{g}^{-\frac{1}{2}}}\sum_{i=1}^{\mu}w_{i}{\bf y}_{i:\lambda}^{g+1},
\end{gather} 
where $c_{\sigma}$ is the learning rate for the \textit{conjugate} evolution path.

Using the \textit{conjugate} evolution path, the step size of the next generation is adjusted. This is to ensure a faster convergence and either increase the step size if the steps recorded are pointing in the same direction or decrease the step size if the steps recorded are not converging and move in opposite directions. This is achieved by comparing the length of the \textit{conjugate} evolution path, ${\bf p}_{\sigma}^{g+1}$, with its expected length, $\mathbb{E}||\mathcal{N}({\boldsymbol{0}},{\bf I})||$. The adaptation of the step size can be expressed as \cite{Hansen2016}:
\begin{gather}
  {\sigma}^{g+1} = {\sigma}^{g}\exp{(\frac{c_{\sigma}}{d_{\sigma}}(\frac{||{\bf p}_{\sigma}^{g+1}||}{\mathbb{E}||\mathcal{N}({\boldsymbol{0}},{\bf I})||}-1))},
\end{gather}
where $d_{\sigma}$ is a damping parameter.

The implementation of CMA-ES used for this study is summarised in Algorithm \ref{alg:CMA-ES}.

\begin{algorithm}[t]
    \caption{Pseudocode for CMA Evolution Strategy}
    \label{alg:CMA-ES}
    \DontPrintSemicolon
    \KwIn{$0 < {\bf w} < 1$}
        ${\bf C} \gets {\bf I}$, ${\bf p}_{c} \gets {\boldsymbol{0}}$, ${\bf p}_{\sigma} \gets {\boldsymbol{0}}$, $g \gets 0$, ${\bf m} \gets \frac{1}{2}{\bf K}_{max}$
    
    \While{$g < G$}{
        ${\bf x}^{g} \gets \text{sample\_population}({\bf m}, \sigma, {\bf C})$ \;
        $f({\bf x}^{g}) \gets \text{evaluate\_population}(\text{HIL experiments})$ \;
        ${{\bf x}_{i:\lambda}^{g}} \gets \text{sort\_population}({\bf x}^{g},f({\bf x}^{g}))$ \;
        ${\bf m}^{g+1} \gets \text{update\_mean}({\bf m}^{g}, {\bf w}, {\bf x}_{i:\lambda}^{g}, {\sigma})$ \;
        ${\bf p}_{\sigma}^{g+1} \gets \text{update\_conj\_path}({\bf p}_{\sigma}^{g}, {\bf C}, {\bf m}^{g}, {\bf w}, {\bf x}_{i:\lambda}^{g}, {\sigma})$ \;
        ${\sigma}^{g+1} \gets \text{update\_step\_size}({\sigma}^{g},{\bf p}_{\sigma}^{g+1})$ \;
        ${\bf p}_{c}^{g+1} \gets \text{update\_cum\_path}({\bf p}_{c}^{g}, {\bf m}^{g}, {\bf w}, {\bf x}_{i:\lambda}^{g}, {\sigma})$ \;
        ${\bf C}^{g+1} \gets \text{update\_C}({\bf C}^{g}, {\bf p}_{c}^{g+1}, {\bf m}^{g}, {\bf w}, {\bf x}_{i:\lambda}^{g}, {\sigma})$ \;
        ${\bf S} \gets \{{\bf S}; [{\bf x}^{g}, f({\bf x}^{g})]\}$ \;
        $g \gets g+1$ \;
    }
    ${\bf x}^{*} \gets {\bf S}$ \text{such that} $f(\bf x)^{*}=\min \bf S$ \;
\end{algorithm}

\section{EXPERIMENTAL VALIDATION}
\begin{figure*}[t]
  \centering
  \includegraphics[width=0.99\linewidth]{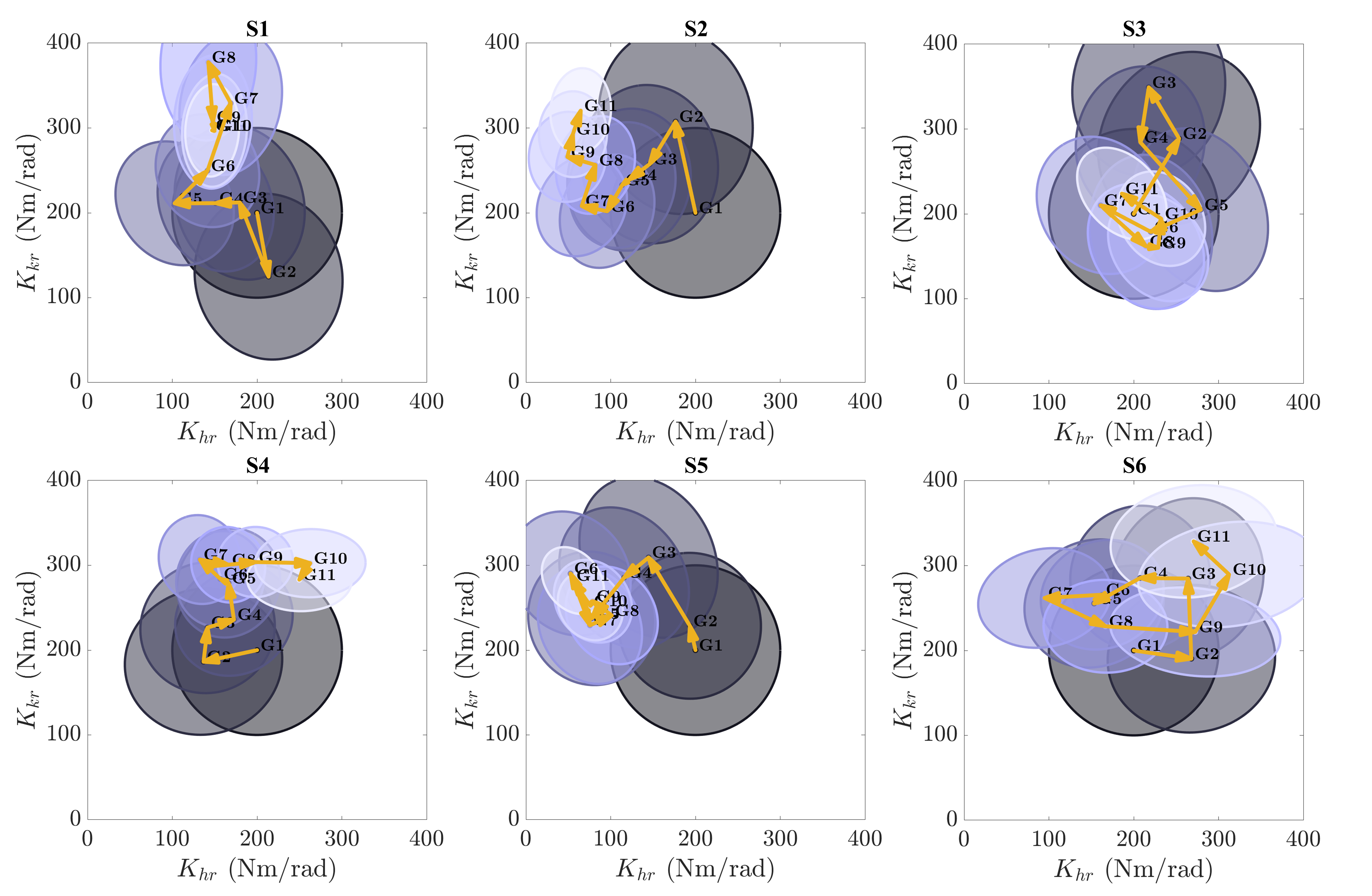}
  \caption[CMA-ES convergence]{Adaptation of covariance matrix and the CMA-ES generation mean, indicate from G1-G11, for all generations and all subjects. Yellow arrows show the CMA-ES step size and direction, and map out the progression of the generation mean.}
  \label{fig:CMAES_Cov}
\end{figure*}
\subsection{Subjects}
The effectiveness of the HILO was tested on six healthy subjects (age = $30 \pm 4.6$, weight = $73.4 \pm 10.5$kg, 2 females). All participants were first-time users of a wearable robot. The experiment pipeline was approved by the University of Edinburgh, School of Informatics Ethics Committee (ID 2021/46920) and the participants provided written consent.

\subsection{Hardware}
The instrumented treadmill M-Gait (Motek Medical, Netherlands) was used to enable self-paced gait during the experiment and the exoskeleton Exo-H3 (Technaid, Spain) was used to provide assistance during gait (an upgrade of the version Exo-H2 presented in \cite{Bortole2015}). The exoskeleton's joint position sensors were used to record the joint angles of the legs and provide real-time visual feedback to the user (Fig. \ref{fig:HILOExperiments}a). Simulink Desktop Real Time was used for the real-time control of the exoskeleton at 100 Hz.  

\subsection{Experimental Setup}
The experiments involved assisted gait training at different levels of exoskeleton stiffness. The participants were fitted with the exoskeleton and were asked to walk on the treadmill at their preferred speed in order to track the reference path as accurately as possible. The recorded kinematics of a healthy subject were used as the reference path, and the path was adjusted to a path of a less pronounce loading response. Adjustments to the reference path were also made to increase comfort for each subject.

Prior to optimisation, a training period was included to familiarise the subjects with the task and the visual feedback. At all times, real-time visual feedback was provided to allow the participants to compare their kinematics to the reference, and inform them about the remaining duration of the experiment and their performance. During optimisation, a new exoskeleton stiffness was tested every minute and the performance of the subjects was measured as described by equation \ref{eq:ObjFunHILO}. To reduce bias from the changing exoskeleton stiffness, the first 15 seconds of each trial were discarded. One generation of the CMA-ES included 7 sample points, making up a 7-minute bout. After each 7-minute bout, a 5-minute break was allowed to reduce bias from fatigue. A total of 5 generations were performed per day, resulting in a total of 10 generations, or 70 sampled points, over the two-day experiment. At the end of each day three rounds of 3-minute validation trials were carried out in a randomised order to evaluate the effectiveness of the HILO. These included a trial with no assistance, a trial with the baseline stiffness (defined as ${\bf K}_{0}=200 \text{Nm/rad}$), a trial with the best identified stiffness, and a trial with the last mean stiffness obtained from the last CMA-ES generation (Figure \ref{fig:HILOExperiments}b). Lastly, a trial with no assistance was carried out.

To reduce the risk of overwhelming the participants with sensory information, subjects were asked to carry out the experiment on only one of the two legs (while the other leg was controlled with a constant low stiffness $K=50$ Nm/rad). The convergence of the HILO and its effect on the performance of the participants are observed and discussed.   

\subsection{Analysis}
Statistical analysis was carried out using two-way repeated-measures ANOVA with controller type (baseline, optimised, and last mean) and time (day 1 and day 2) as within-subject factors. Additionally, a one-way repeated measures ANOVA was performed to analyse differences between controller types regardless of time. Shapiro-Wilk’s test of normality was applied to model residuals, and Mauchly’s test was conducted to evaluate sphericity. If the sphericity assumption was violated, epsilon corrections were applied. Following the ANOVA, post-hoc pairwise comparisons were performed using MATLAB’s \textit{multcompare} function with Bonferroni correction to account for multiple comparisons.

\begin{table}[t]
    \centering
    \caption{Hyperparameter values for the controller and optimisation algorithm.}
    \label{tab:hyperparams}
    \begin{tabular}{l c c}
        \toprule
        \multicolumn{3}{c}{\textbf{Controller Hyperparameters}} \\
        \midrule
        Max Exoskeleton assistance & ${\bf u}_{max}$ & 40Nm \\
        Max expected stiffness & ${\bf K}_{max}$ &  $400Nm/rad$\\
        Max trajectory error & $\Delta {\bf q}_{max}$ & $2\textdegree$ \\
        Baseline stiffness & ${\bf K}_{0}$ &  $200Nm/rad$\\
        Critical damping & $c_r$ & 10 \\
        Effort cost weight & $w_1$ & 3 \\
        Tracking cost weight & $w_2$ & 1 \\
        Stiffness cost weight & $w_3$ & 0.1 \\
        \midrule
        \multicolumn{3}{c}{\textbf{CMA-ES Hyperparameters}} \\
        \midrule
        Sample points & $\lambda$ & 7 \\
        Step size & $\sigma_{0}$ & 150 \\
        LR - mean & $c_m$ & 1 \\
        LR - rank-one covariance & $c_1$ & 0.15 \\
        LR - rank-$\mu$ covariance & $c_\mu$ & 0.058 \\
        LR - rank-$\sigma$ covariance & $c_\sigma$ & 0.62 \\
        \bottomrule
    \end{tabular}
\end{table}

\section{RESULTS}

Figure \ref{fig:CMAES_Cov} shows the adaptation of the CMA-ES for all subjects. A continuous adaptation of the CMA-ES can be seen. In most cases this led to an assistive controller which included a higher stiffness at the knee joint compared to both the hip joint and the baseline stiffness. This suggests that despite the variability in gait between and within individuals, for the majority of the 2-day trial the participants were able to follow the reference path more accurately when the stiffness of the knee joint was higher than the baseline stiffness and the hip stiffness. It can also be seen that some participants were more consistent with their performance than others. This is revealed by the direction in which the mean points of the CMA-ES progress. For example, the progression of the generated mean points for subjects S1, S2, S4 and S5 appears to be smoother and more gradual, than subjects S3 and S6. This also led to a gradually decreasing covariance which is less evident in the results obtained for subjects S3 and S6. 

\begin{figure}[t]
  \centering
  \includegraphics[width=0.99\linewidth]{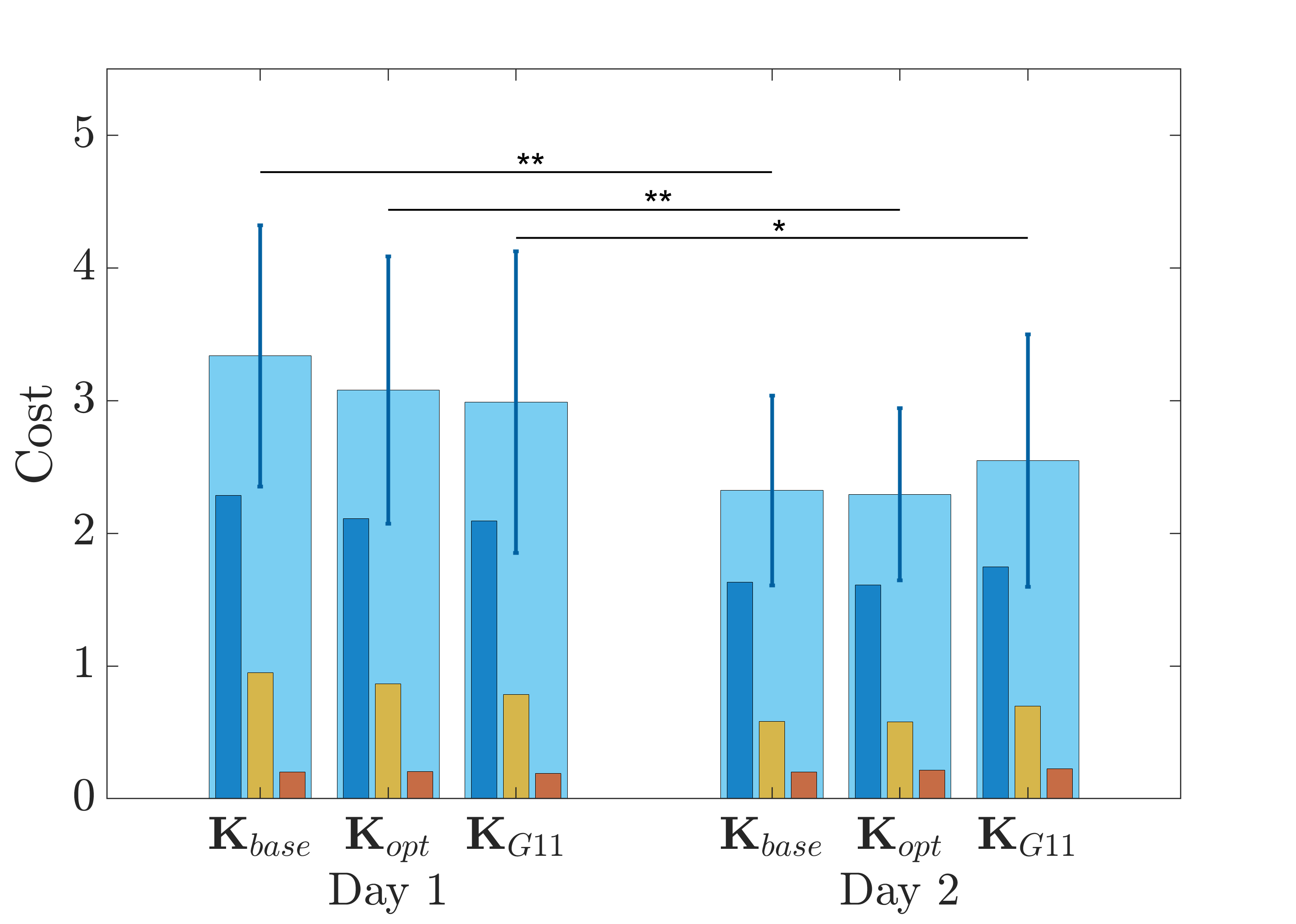}
  \caption[Validation of HILO]{Results from validation trials with the baseline stiffness, ${\bf K}_{base}$, the optimised stiffness, ${\bf K}_{opt}$, and the last mean stiffness, ${\bf K}_{G11}$, for day 1 and day 2 separately. Error bars show the standard deviation of cost (*P$<$0.05, **P$<$0.01).}
  \label{fig:HILO_Validation_Day1_Day2_Separately}
\end{figure}

\begin{figure}[t!]
  \centering
  \includegraphics[width=0.815\linewidth]{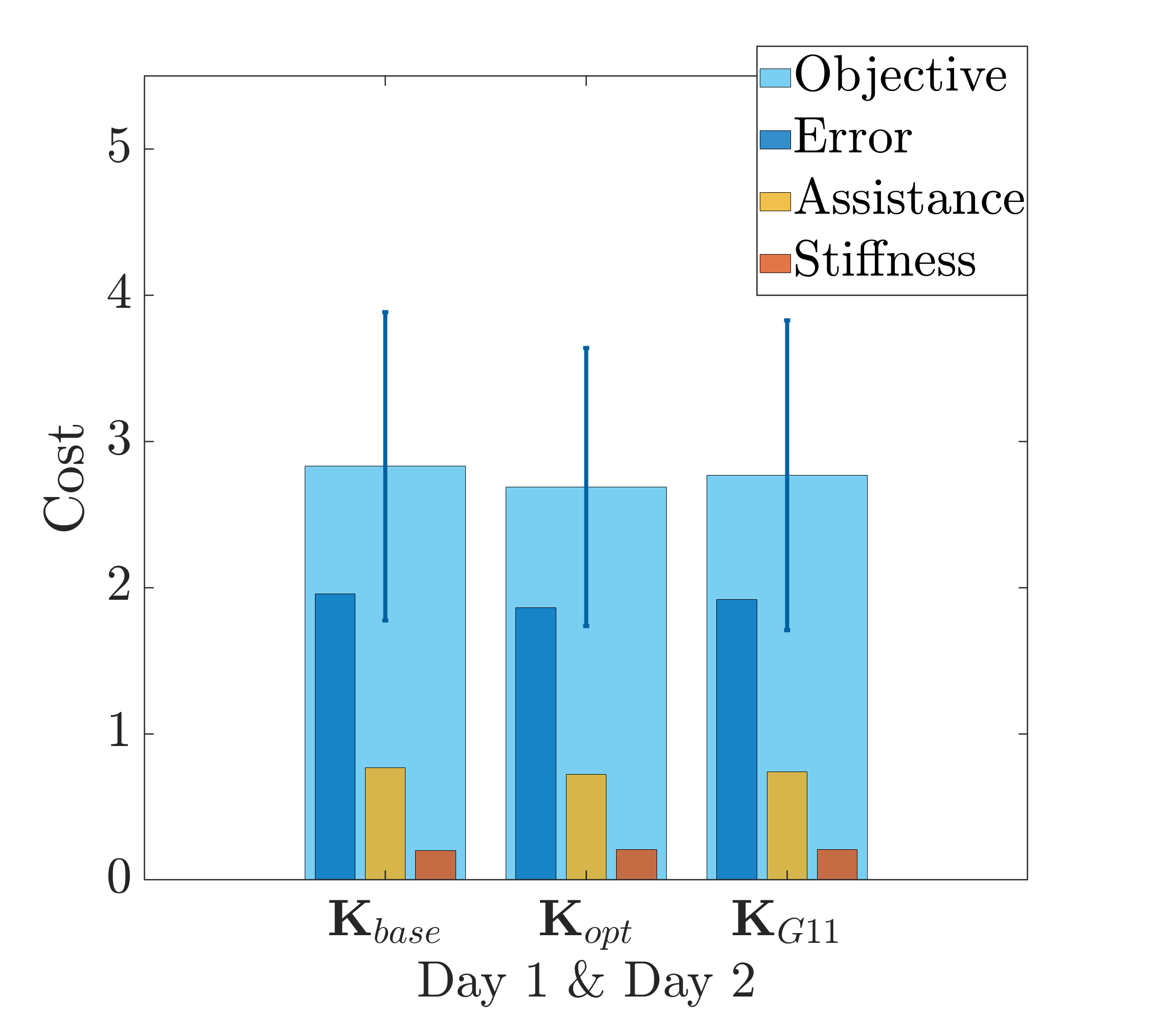}
  \caption[Validation of HILO]{Results from validations trials with the baseline stiffness, ${\bf K}_{base}$, the optimised stiffness, ${\bf K}_{opt}$, and the last mean stiffness, ${\bf K}_{G11}$, for day 1 and day 2 combined. Error bars show the standard deviation of cost.}
  \label{fig:HILO_Validation_Day1_Day2_Together}
\end{figure}

In Figures \ref{fig:HILO_Validation_Day1_Day2_Separately}-\ref{fig:HILO_Validation_Day1_Day2_Together} the results from the validation trials are presented. It can be seen that the performance of the subjects has significantly improved from day 1 to day 2 with all controllers (Figure \ref{fig:HILO_Validation_Day1_Day2_Separately}). However, no significant differences were observed between groups when the performance of the subjects was compared when using the baseline stiffness, the optimised stiffness or the stiffness at the last mean point of the CMA-ES (Figure \ref{fig:HILO_Validation_Day1_Day2_Together}). High performance variability was observed both between and within subjects, which may have detracted from the benefits due to the adjusted controller stiffness.

\section{DISCUSSION}

Here, an implementation of HILO is proposed for the personalisation of robot-assisted gait training. For this purpose the CMA-ES is used and a continuous optimisation experimental protocol is suggested to allow enough time for the optimiser to converge over multiple days. This aligns with the needs of gait training, where gait trials may need to be interrupted for rest and safety. To the best of our knowledge, this is the first time HILO has been tested for the personalisation of assistive controllers in gait rehabilitation and for robots with multiple DOFs. It is hypothesised that with the human in the loop, the robot controller can be iteratively optimised based on the user's performance in order to provide assistance as needed. This could in turn allow ambulatory patients to benefit from robot-assisted gait training in order to increase independence and gait efficiency.

Results from six healthy subjects are obtained indicating that the CMA-ES can adapt the stiffness of the robot controller based on the performance of the subjects to personalise gait training. However, conclusive results regarding the effectiveness of HILO in providing assistance as needed could not be obtained. One major challenge in HILO, and in the personalisation of closed-loop robotic controllers, is intrapersonal variability. Variability within individuals may be both time-dependent and random, which poses a significant challenge in the design of personalised interventions. For an adaptive control algorithm to prove effective, the time-dependent changes in human behaviour need to be captured and the benefits of adaptation need to outweigh any effects of unpredicted behaviour variability. This appears to be especially hard when the adaptation of rule-based controllers is considered, since such controllers are expected to be effective within a range of control parameters.

A key factor contributing to this variability is the dynamic interaction between human and robot, where both systems continuously adapt to each other. On the one hand, adjustments to the stiffness of the robot are carried out to provide support to the human in an optimal way, but on the other hand, the human behaviour concurrently changes to utilise this stiffness perhaps for a different subject-specific objective (e.g. comfort or metabolic efficiency). This co-adaptation, influenced by factors such as fatigue, comfort, motor learning, and concentration, introduces additional complexity and bias into the optimisation process that may outweigh the intended benefits of robot adaptation. In healthy individuals, this voluntary adaptation is likely more pronounced, as they have greater control over their lower limbs and can actively respond to changes in assistance. Conversely, neurological patients with impaired motor function may exhibit less voluntary adaptation, potentially reducing optimisation bias due to co-adaptation. However, this advantage is counterbalanced by other challenges. Neurological patients are more prone to fatigue, unintentional motor commands, and increased variability in movement patterns, all of which can introduce noise into the optimisation process and hinder convergence.

\section{CONCLUSION}

Our work contributes to understanding the limitations of current personalisation approaches in robot-assisted gait rehabilitation and identifies key challenges for effective implementation of human-in-the-loop optimisation in this domain. Our results from six individuals suggest that while the CMA-ES can continuously adapt the stiffness parameters over time, the impact of these adaptations on performance may be masked by the high degree of human variability and co-adaptation. Future studies can focus on refining optimisation protocols to better account for human behavioural variability in order to improve the effectiveness of HILO. This may include more noise-resilient algorithms or hybrid approaches that combine HILO with therapist input or model-based constraints which may be able to more accurately account for the human-robot co-adaptation. Tailoring experimental protocols to patient-specific needs, whether by adjusting trial durations, refining performance metrics, or mitigating fatigue-related biases could also prove beneficial.
Moreover, identifying combinations of rehabilitation tasks and assistive controllers where personalisation is likely to yield substantial gains is critical, particularly in cases where controller performance is highly sensitive to parameter selection, as opposed to strategies that are inherently more robust across a wider range of parameters. Finally, investigating HILO in clinical populations, such as individuals post-stroke where voluntary adaptation is more constrained, could offer clearer insight into its practical benefits and play a key role in realising the full potential of personalisation to improve clinical outcomes and accelerate recovery in robot-assisted rehabilitation.

\addtolength{\textheight}{-0cm}   









\bibliographystyle{IEEEtran}

\end{document}